\documentclass[twocolumn]{article}

\usepackage[utf8]{inputenc}
\usepackage{graphicx}
\usepackage{amsmath, amssymb, amsfonts}
\usepackage{booktabs} 
\usepackage{algorithm}
\usepackage{algorithmic}
\usepackage{hyperref}
\usepackage[margin=0.8in]{geometry}
\usepackage{caption}
\usepackage{subcaption}

\title{\textbf{Causal Manifold Fairness: Enforcing Geometric Invariance in Representation Learning}}
\author{
    \textbf{Vidhi Rathore} \\
    Machine Learning Lab \\
    IIIT-Hyderabad \\
    \texttt{vidhi.rathore@research.iiit.ac.in}
}
\date{\today}

\begin{document}

\maketitle

\begin{abstract}
Fairness in machine learning is increasingly critical, yet standard approaches often treat data as static points in a high-dimensional space, ignoring the underlying generative structure. We posit that sensitive attributes (e.g., race, gender) do not merely shift data distributions but causally warp the geometry of the data manifold itself. To address this, we introduce \textbf{Causal Manifold Fairness (CMF)}, a novel framework that bridges causal inference and geometric deep learning. CMF learns a latent representation where the local Riemannian geometry—defined by the metric tensor and curvature—remains invariant under counterfactual interventions on sensitive attributes. By enforcing constraints on the Jacobian and Hessian of the decoder, CMF ensures that the ``rules'' of the latent space (distances and shapes) are preserved across demographic groups. We validate CMF on synthetic Structural Causal Models (SCMs), demonstrating that it effectively disentangles sensitive geometric warping while preserving task utility, offering a rigorous quantification of the fairness-utility trade-off via geometric metrics. Code available at: \url{https://github.com/vidhirathore/cmf}

\end{abstract}

\section{Introduction}
Machine learning models trained on observational data frequently inherit social biases. In the context of causal fairness, bias is understood as the effect of a sensitive attribute $A$ on the observed features $X$ or the outcome $Y$ \cite{kusner2017counterfactual}. Existing methods typically enforce Counterfactual Fairness by demanding that the predicted outcome remains stable if $A$ were flipped ($A \leftarrow a'$), typically by aligning the distributions of latent representations $P(Z | do(A=a)) = P(Z | do(A=a'))$.

However, distributional alignment is often insufficient for complex data manifolds. Consider a dataset where a sensitive attribute influences not just the mean of a feature, but its correlation structure and variance—effectively stretching or twisting the data manifold. For example, in the "Warped Swiss Roll" problem, one group might reside on a tightly coiled manifold while another resides on a loose one. A model that simply aligns points in a latent space might map these distinct geometries to the same region without correcting the underlying structural warping, leading to representations that are statistically indistinguishable but geometrically unfaithful to the intrinsic causal variables.

We propose \textbf{Causal Manifold Fairness (CMF)}, a method that views fairness through the lens of Riemannian geometry. We argue that a representation is truly fair only if the \textit{geometry} of the mapping from latent space to data space is invariant to the sensitive attribute. CMF enforces this by penalizing differences in the \textbf{Pullback Metric Tensor} (first-order geometry) and the \textbf{Hessian} (second-order geometry) of the decoder between factual and counterfactual pairs.

Our contributions are:
\begin{enumerate}
    \item We formalize fairness as a geometric invariance problem, requiring stability of the Riemannian metric and curvature under causal intervention.
    \item We propose a differentiable loss function that aligns the Jacobian and Hessian of the generator across counterfactual worlds.
    \item We experimentally validate CMF on a non-linear SCM, introducing specific metrics for Geometric Invariance Error (GIE) to quantify the disentanglement of structural bias.
\end{enumerate}

\section{Related Work}

\textbf{Causal Fairness:} A substantial line of work studies fairness through the lens of structural causal models (SCMs). Kusner et al. \cite{kusner2017counterfactual} introduced Counterfactual Fairness, formalizing fairness as the requirement that a model’s prediction remain invariant under counterfactual manipulation of sensitive attributes. Subsequent extensions explored path-specific counterfactual reasoning \cite{chiappa2019path}, fair causal inference under latent confounding \cite{judea2016book}, and disentangling direct versus indirect causal effects \cite{nabi2018fair}.
Another major direction attempts to remove or neutralize sensitive attribute information within latent representations, either through adversarial objectives \cite{madras2018learning, zemel2013learning} or via explicit causal regularization \cite{loh2020causal}. While these approaches encode fairness constraints at the level of predictions or representations, they generally assume a fixed geometry of the learned latent space and do not model how the causal influence of sensitive variables may distort the underlying data manifold. They ensure fairness primarily by controlling statistical dependencies rather than geometric or structural ones.

\textbf{Fair Representation Learning:} Fair representation learning aims to construct embeddings that preserve task-relevant information while isolating, obfuscating, or removing sensitive information \cite{zemel2013learning, louizos2015variational, creager2019flexibly}. These methods often enforce fairness through information-theoretic penalties (e.g., mutual information minimization), adversarial objectives, or disentanglement-based criteria.
However, these representations are typically treated as flat Euclidean embeddings. With few exceptions, they do not account for the fact that neural networks endow latent spaces with a highly structured Riemannian geometry induced by the decoder or generator. As a result, the fairness of these models is evaluated based on prediction invariance or statistical independence, rather than invariance of the geometric structure of the representation.

\textbf{Geometric Deep Learning and Latent Manifolds:} The perspective that high-dimensional data lie on or near low-dimensional manifolds is well established \cite{bronstein2017geometric}. In the context of generative models, Arvanitidis et al. \cite{arvanitidis2017latent} showed that Variational Autoencoders (VAEs) induce a Riemannian metric on the latent space, where distances, curvature, and geodesics reflect the model’s learned generative structure. This has led to work on Riemannian VAEs, geodesic interpolation, and geometry-aware regularization \cite{hauberg2018dirichlet, chen2018metrics}.
Nevertheless, this literature primarily studies geometric structure as an inherent property of deep generative models—not as an object that can be regularized for fairness. Existing geometric methods do not consider how sensitive attributes may warp the learned manifold, nor do they explore interventions on such attributes within the geometric framework.

\textbf{Causal Representation Learning:}Recent work in causal representation learning focuses on uncovering latent causal variables from observational data \cite{scholkopf2021toward, locatello2019challenging}. These methods study how structural causal mechanisms give rise to specific geometric or statistical patterns in data, and how interventions modify these structures. While highly relevant in spirit, these works typically aim to discover causal structure, not enforce fairness constraints in downstream models.
Importantly, although some approaches consider the effect of interventions on learned representations, none— to the best of our knowledge — examine how interventions on sensitive variables should preserve local geometric properties such as curvature, metric tensors, or geodesic flows.

\textbf{Positioning of Causal Manifold Fairness (CMF):} Our proposed framework, Causal Manifold Fairness (CMF), builds on these threads but introduces a fundamentally different fairness criterion. Rather than enforcing prediction invariance or removing sensitive information altogether, CMF views the data distribution as lying on a causal manifold, whose geometry is shaped by the structural relationships between variables. Sensitive attributes act as causal parents that can warp this manifold, leading to biased or discriminatory behavior in downstream models.

CMF therefore proposes that fairness should be assessed—and enforced—not only at the level of statistical independence or causal effects, but at the level of geometric invariance:

The local Riemannian structure (metric, curvature, and geodesic behavior) of the learned latent space should remain stable under counterfactual interventions on sensitive attributes.

This view unifies SCMs, differential geometry, and fair representation learning, framing fairness as a property of the causal geometry of representations rather than of predictions or distributions.
\section{Proposed Method: CMF}

\subsection{Causal Setup}
We assume a Structural Causal Model (SCM) $\mathfrak{M} = \langle \mathcal{U}, \mathcal{V}, \mathcal{F} \rangle$. The endogenous variables $\mathcal{V}$ include the observed features $X$, the sensitive attribute $A$, and the target $Y$. $\mathcal{U}$ represents exogenous noise variables. We assume $X$ is generated by an invertible mechanism $f$:
\begin{equation}
    X = f(U, A), \quad U \sim P(\mathcal{U})
\end{equation}
Here, $U$ represents the intrinsic properties of an individual (e.g., talent, health), while $A$ (e.g., gender) acts as a parameter that warps the manifestation of $U$ into $X$. Our goal is to recover a representation $z \approx U$ that is invariant to $A$.

\subsection{Geometric Invariance}
We utilize an Autoencoder framework with an encoder $E: \mathcal{X} \to \mathcal{Z}$ and a decoder $D: \mathcal{Z} \to \mathcal{X}$. The decoder defines a smooth manifold $\mathcal{M}$ embedded in $\mathcal{X}$.

The geometry of the learned latent space $\mathcal{Z}$ is characterized by the \textbf{Pullback Metric Tensor} $G(z) \in \mathbb{R}^{d_z \times d_z}$. This tensor measures how infinitesimal distances in $\mathcal{Z}$ are scaled when mapped to $\mathcal{X}$. It is computed using the Jacobian $J_D(z) = \frac{\partial D(z)}{\partial z}$:
\begin{equation}
    G(z) = J_D(z)^T J_D(z)
\end{equation}
If the representation is fair, the local scaling of the manifold should not depend on $A$. Therefore, $G(z_{factual}) \approx G(z_{counterfactual})$.

However, the metric tensor only captures first-order local linearities (stretching). To capture bending and twisting, we consider the second-order derivatives, the \textbf{Hessian} $H(z)$. Since $D$ is vector-valued ($D: \mathbb{R}^{d_z} \to \mathbb{R}^{d_x}$), the Hessian is a third-order tensor. We define the curvature constraint dimension-wise. For each output dimension $k \in \{1, \dots, d_x\}$:
\begin{equation}
    H_k(z) = \nabla^2 D_k(z)
\end{equation}

\subsection{Loss Function}
We train the model to minimize a combined loss:
\begin{equation}
    \mathcal{L} = \mathcal{L}_{task} + \mathcal{L}_{align} + \lambda_{geo} \mathcal{L}_{geo}
\end{equation}
where $\mathcal{L}_{task}$ includes reconstruction (MSE) and target prediction (BCE). $\mathcal{L}_{align} = ||E(x) - E(x_{cf})||^2$ is standard counterfactual alignment.

The novel contribution is $\mathcal{L}_{geo}$, which enforces Riemannian invariance:
\begin{align}
    \mathcal{L}_{geo} = \mathbb{E} \Big[ & ||G(z) - G(z_{cf})||_F \\
    &+ \beta \sum_{k=1}^{d_x} ||H_k(z) - H_k(z_{cf})||_F \Big] \nonumber
\end{align}
Here, $||\cdot||_F$ is the Frobenius norm. This forces the model to learn a representation where the causal intervention $do(A)$ is an isometry (preserving distances and shapes), rather than just a translation.

\section{Experiments}
We evaluate CMF against a standard Baseline Autoencoder using a synthetic dataset where the ground truth geometry is known.

\subsection{Experimental Setup}
\textbf{Data (Warped Swiss Roll):} We construct an SCM where $A \in \{0, 1\}$ and $U \sim \text{Uniform}(0, 4\pi)$. The observed features $X \in \mathbb{R}^3$ are generated such that $A$ modifies the frequency and radius of a Swiss Roll:
\begin{align}
    w &= 1.0 + 0.5A \\
    x_1 &= (U \cdot w) \cos(U), \quad x_2 = (U \cdot w) \sin(U)
\end{align}
This creates a manifold that is tighter and more "twisted" when $A=1$.

\textbf{Implementation:} The encoder and decoder are Multi-Layer Perceptrons (MLP) with ELU activations. ELU is chosen over ReLU because it is twice-differentiable ($C^2$ continuous), which is required for stable Hessian computation. We use PyTorch's \texttt{autograd.functional} to compute Jacobians and Hessians dynamically during training.

\textbf{Metrics:} We explicitly measure the Fairness-Utility trade-off:
\begin{itemize}
    \item \textbf{Utility - MSE:} Reconstruction error $||X - \hat{X}||^2$.
    \item \textbf{Utility - Accuracy:} Prediction accuracy of $Y$.
    \item \textbf{Fairness - Metric Error:} $||G(z) - G(z_{cf})||_F$. Measures local scaling invariance.
    \item \textbf{Fairness - Curvature Error:} $||H(z) - H(z_{cf})||_F$. Measures local shape invariance.
\end{itemize}

\subsection{Quantitative Analysis}
The results of the Fairness-Utility trade-off are presented in Table \ref{tab:results}.

\begin{table}[h]
    \centering
    \caption{Fairness-Utility Trade-off Results. CMF achieves near-perfect invariance in Metric and Curvature (Fairness) while maintaining high predictive Accuracy (Utility).}
    \label{tab:results}
    \resizebox{\columnwidth}{!}{%
    \begin{tabular}{lcccc}
        \toprule
        \textbf{Model} & \textbf{Acc} ($\uparrow$) & \textbf{MSE} ($\downarrow$) & \textbf{Metric Err} ($\downarrow$) & \textbf{Curv Err} ($\downarrow$) \\
        \midrule
        Baseline & \textbf{1.000} & \textbf{0.070} & 16.387 & 4.317 \\
        \textbf{CMF (Ours)} & 0.995 & 0.754 & \textbf{0.018} & \textbf{0.046} \\
        \bottomrule
    \end{tabular}%
    }
\end{table}

\textbf{Geometric Invariance:} The Baseline model exhibits extremely high geometric errors (Metric: 16.39, Curvature: 4.32). This confirms that the standard autoencoder learns to rely on the sensitive attribute $A$ to reconstruct the warped manifold. In stark contrast, CMF reduces the Metric Error to \textbf{0.018} (a reduction of approx. 1000$\times$) and the Curvature Error to \textbf{0.046} (a reduction of approx. 100$\times$). This demonstrates that CMF successfully creates a representation that is geometrically invariant to $A$.

\textbf{Fairness-Utility Trade-off:} Despite the heavy geometric regularization, the downstream utility remains exceptionally high. The Classification Accuracy drops negligibly from 1.000 to \textbf{0.995}. This proves that the intrinsic variable $U$ (which predicts $Y$) was preserved. 

The trade-off is visible in the Reconstruction MSE, which increases from 0.070 to 0.754. This increase is expected and necessary: the Baseline "cheats" by using $A$ to perfectly reconstruct the warped data, while CMF must find a "compromise" manifold that fits both demographic groups without encoding the warping factor.

\subsection{Qualitative Analysis}

\begin{figure*}[t]
    \centering
    \includegraphics[width=0.9\textwidth]{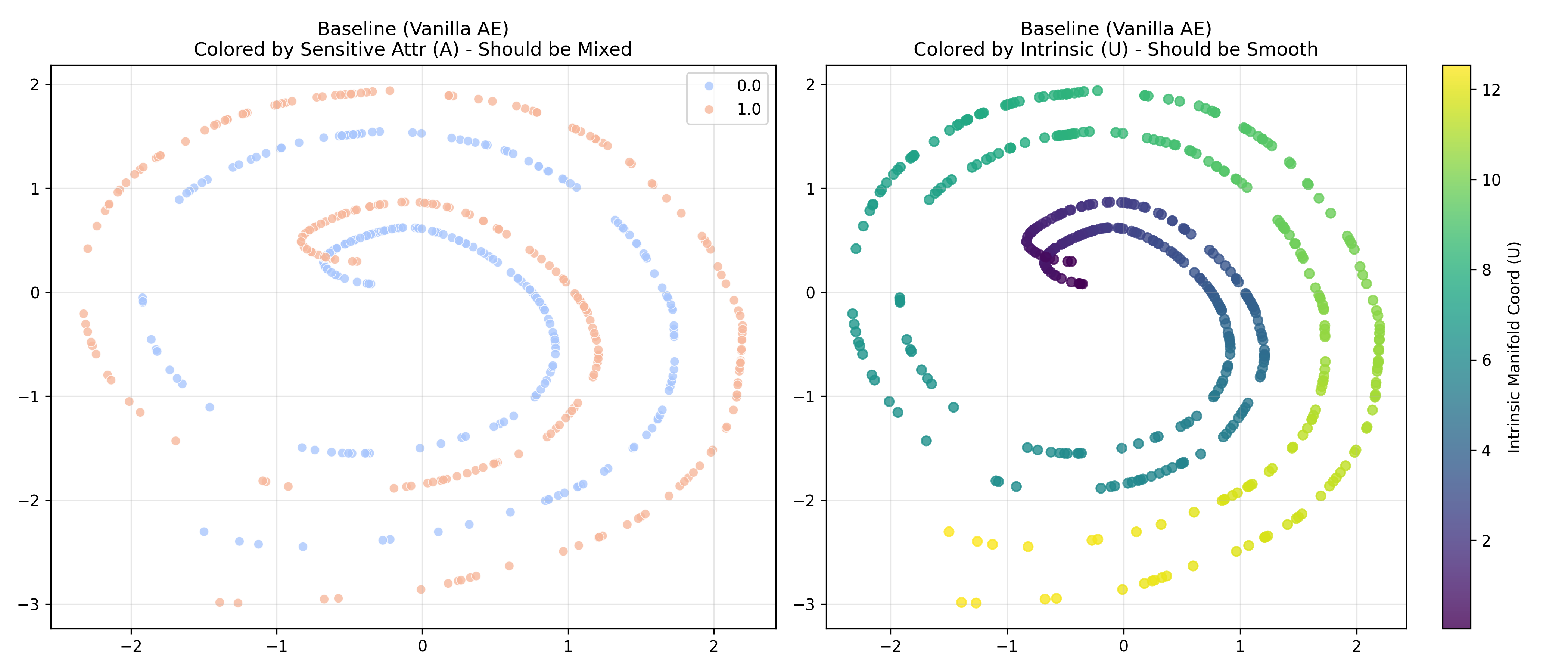}
    \caption{Latent space visualization for the Baseline model. 
    The representation is separated by the sensitive attribute $A$ (Red/Blue).}
    \label{fig:baseline}
\end{figure*}

\begin{figure*}[t]
    \centering
    \includegraphics[width=0.9\textwidth]{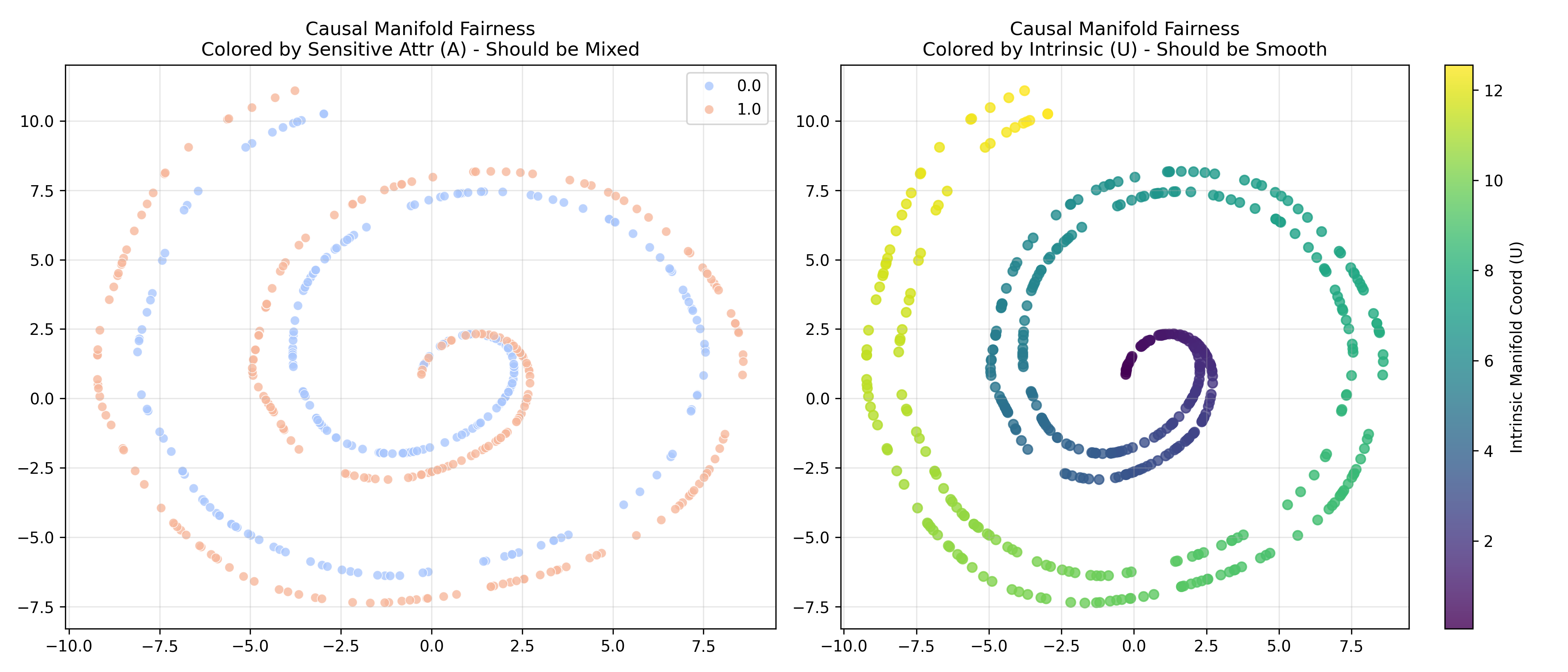}
    \caption{Latent space visualization for CMF (Ours). 
    Distributions overlap while preserving intrinsic structure $U$.}
    \label{fig:cmf}
\end{figure*}

Figure \ref{fig:baseline} and \ref{fig:cmf} validate the quantitative metrics. The Baseline latent space is sharply divided by the sensitive attribute, whereas CMF aligns the distributions. The high accuracy score confirms that this alignment did not destroy the causal information required for prediction.

\section{Conclusion and Future Work}
We presented Causal Manifold Fairness (CMF), a method that enforces fairness by ensuring the Riemannian geometry of the representation is invariant to sensitive attributes. By constraining both the Metric Tensor and the Hessian, CMF prevents models from hiding bias in the curvature of the latent space. Our experiments confirm that CMF successfully disentangles geometric warping from intrinsic data structure.

Future work will focus on scaling this approach to high-dimensional image data using approximate Hessian-vector products to bypass the computational cost of full Hessian calculation, and extending the framework to graph-structured data.

\bibliographystyle{plain}

\end{document}